\documentclass[twocolumn]{article}
\usepackage{amsmath}
\usepackage{graphicx}

\usepackage{amsmath}
\usepackage{amsfonts}
\usepackage{amssymb}
\usepackage{url}
\usepackage{bm}
\usepackage{bbm}
\usepackage{float}
\usepackage{algorithm}
\usepackage{algpseudocode}
\usepackage{multirow}
\usepackage{bbm}
\usepackage[table,xcdraw]{xcolor}

\makeatletter
\newcommand{\algmargin}{\the\ALG@thistlm}
\makeatother
\newlength{\whilewidth}
\algdef{SE}[parWHILE]{parWhile}{EndparWhile}[1]
{\parbox[t]{\dimexpr\linewidth-\algmargin}{%
		\hangindent\whilewidth\strut\algorithmicwhile\ #1\ \algorithmicdo\strut}}{\algorithmicend\ \algorithmicwhile}%
\algnewcommand{\parState}[1]{\State%
	\parbox[t]{\dimexpr\linewidth-\algmargin}{\strut #1\strut}}

\def\L{{\cal L}}

\DeclareMathOperator*{\argmax}{argmax}
\newcommand{\argmaxA}{\mathop{\mathrm{argmax}}\limits}

\title{END-TO-END FEEDBACK LOSS IN SPEECH CHAIN FRAMEWORK VIA STRAIGHT-THROUGH ESTIMATOR}
\author{Andros Tjandra$^{1,2}$, Sakriani Sakti$^{1,2}$, Satoshi Nakamura$^{1,2}$ \\ \\ $^1$Nara Institute of Science and Technology, Japan\\
	$^2$RIKEN, Center for Advanced Intelligence Project AIP, Japan\\ \texttt{\{andros.tjandra.ai6,ssakti,s-nakamura\}@is.naist.jp}}
\date{}

\begin{document}
	
	\maketitle
	
	\begin{abstract}
		The speech chain mechanism integrates automatic speech recognition (ASR) and text-to-speech synthesis (TTS) modules into a single cycle during training. In our previous work, we applied a speech chain mechanism as a semi-supervised learning. It provides the ability for ASR and TTS to assist each other when they receive unpaired data and let them infer the missing pair and optimize the model with reconstruction loss. If we only have speech without transcription, ASR generates the most likely transcription from the speech data, and then TTS uses the generated transcription to reconstruct the original speech features. However, in previous papers, we just limited our back-propagation to the closest module, which is the TTS part. One reason is that back-propagating the error through the ASR is challenging due to the output of the ASR are discrete tokens, creating non-differentiability between the TTS and ASR. In this paper, we address this problem and describe how to thoroughly train a speech chain end-to-end for reconstruction loss using a straight-through estimator (ST). Experimental results revealed that, with sampling from ST-Gumbel-Softmax, we were able to update ASR parameters and improve the ASR performances by 11\% relative CER reduction compared to the baseline.
	\end{abstract}
	
	\section{Introduction}
	
	A speech chain \cite{denes1993speech} is a viewpoint that describes the speech communication process in which the speaker produces words and generates speech sound waves, transmits the speech waveform through a medium (i.e., air), and creates a speech perception process in a listeners auditory system to perceive what was said. The hearing process is critical, not only for the listener but also for the speaker herself.  By simultaneously listening and speaking, the speaker can monitor her volume, articulation, and the general comprehensibility of her speech.
	Based on those observations, we simulated the speech chain mechanism by coupling ASR and TTS and formed a machine speech chain \cite{tjandra2017speechchain, tjandra2018machinespeech}, so that the machine can learn, not only to listen (by way of ASR) or speak (by way of TTS) but also listen while speaking.
	
\begin{figure*}[h]
		\centering
		\includegraphics[width=0.9\linewidth]{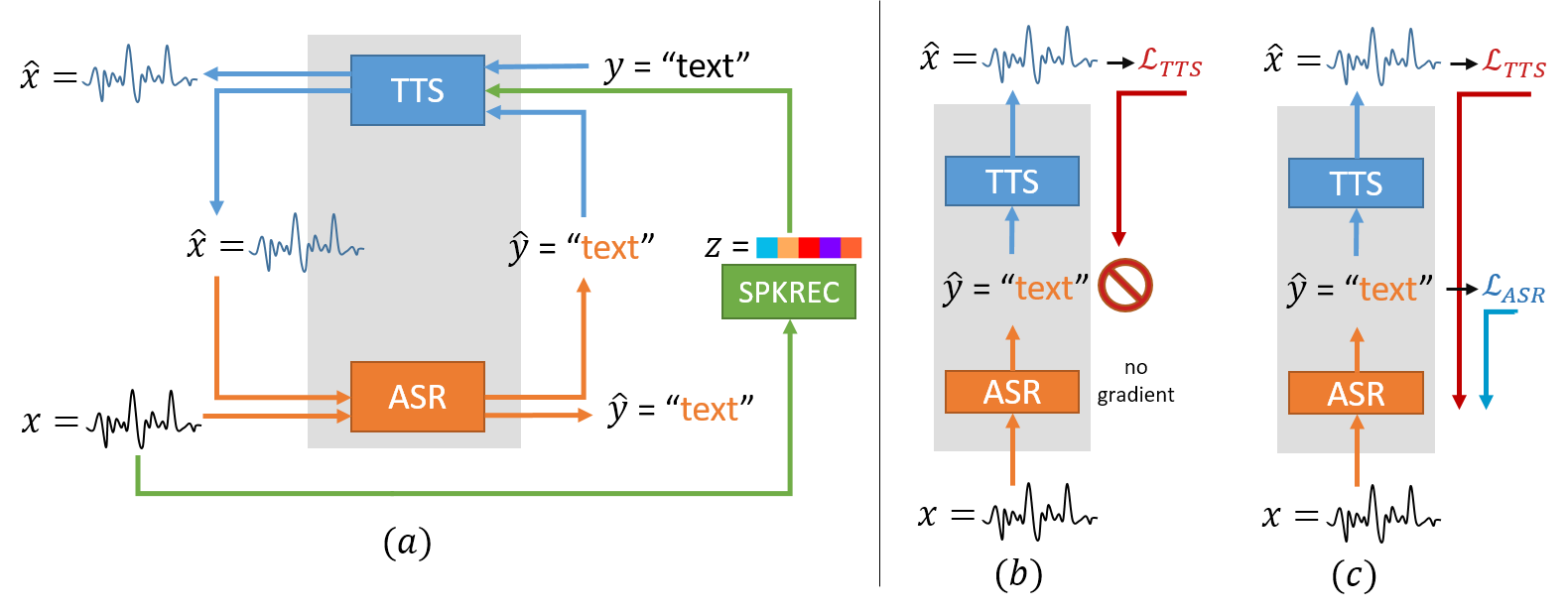}
		
		\caption{a) Multispeaker machine speech chain mechanism; b) Baseline (\cite{tjandra2018machinespeech}): feedback loss from TTS is only backpropagated through the TTS module, and the ASR module is not updated because variable $\hat{y}$ is non-differentiable; c) \textbf{Proposal:} feedback loss from TTS is backpropagated through discrete variable $\hat{y}$, and ASR modules are updated based on the estimated gradient from the TTS module by a straight-through estimator. }
		\label{fig:speech_chain_general}
	\end{figure*}

	In our previous paper \cite{tjandra2017speechchain}, we utilized the speech chain idea for semi-supervised learning using paired and unpaired data. First, we pretrained both ASR and TTS with a small amount of paired speech and text data. Then, we subsequently used both the pretrained modules to complete the missing pair from the unpaired data. For example, if we only have speech without transcription, ASR generates the most likely transcription from the speech data with greedy or beam-search decoding, and TTS uses the generated transcription to reconstruct the original speech features. In this case, we trained the TTS module with the reconstruction loss. For the reverse case, if we only have text without any corresponding speech, TTS generates speech, whose features ASR uses to reconstruct the original text. In this case, we updated the ASR module with the reconstruction loss. In Fig.~\ref{fig:speech_chain_general}(a), we illustrate a multispeaker speech chain loop between the ASR and TTS modules.

	However, the auditory feedback in a human speech chain happens almost constantly, not only during semi-supervised learning. Furthermore, the close-loop feedback is also done end-to-end. But, to simulate our speech chain mechanism to provide the ability to help each other even during the supervised learning and perform a completely end-to-end feedback reconstruction loss, the main challenge is to utilize TTS to improve our ASR module. One reason is that back-propagating the error from the reconstruction loss through the ASR module is challenging due to the output of the ASR discrete tokens (grapheme or phoneme), creating non-differentiability between the TTS and ASR modules (Fig.~\ref{fig:speech_chain_general}(b)).
	
	We address this problem using a straight-through estimator \cite{bengio2013estimating, hinton2012coursera} to predict the gradient through discrete variables (Fig.~\ref{fig:speech_chain_general}(c)). We mainly focus on describing how to thoroughly train a speech chain end-to-end by adding a reconstruction term from the TTS module and backpropagated the gradient through the ASR. Experimental results revealed that, with teacher-forcing and sampling from Gumbel-Softmax, we were now able to updated ASR parameters and improved the ASR performances significantly by 11\% relative CER reduction compared to the baseline.
	
	%\section{Speech Chain}
	
	\section{Speech Chain and End-to-end Feedback Loss} 
	\label{sec:loss_bprop_tts_asr}
	
	In the speech chain mechanism, given speech features $\mathbf{x} = [x_1,..,x_S]$ (e.g., Mel-spectrogram) and $\mathbf{y} = [y_1,..,y_T]$, we feed the speech to the ASR module, and the ASR decoder generates continuous vector $h^d_t$ step-by-step. To calculate probability vector $\mathbf{p}_{y} = [p_{y_1},..,p_{y_T}]$, we apply the softmax function $p_{y_t} = \texttt{softmax}(h^d_t)$ to decoder output $h^e_t$.
	For each class probability mass in $p_{y_t}$, $p_{y_t}[c]$  was defined as:
	
	\begin{equation}
	p_{y_t}[c] = \frac{\exp(h^d_t[c]/\tau)}{\sum_{i=1}^{C} \exp(h^d_t[i] / \tau)}, \quad \forall c \in [1..C]. \label{eq:softmax}
	\end{equation} 
	Here $C$ is the total number of classes, $h_t^d \in \mathbb{R}^{C}$ are the logits produced by the last decoder layer, and $\tau$ is the temperature parameters. Setting temperature $\tau$ using a larger value $(\tau > 1)$ produces a smoother probability mass over classes \cite{hinton2015distilling}.
	
	For the generation process, we generally have two different methods:
	\begin{enumerate}
		\item Conditional generation given ground-truth (teacher-forcing): \\
		If we have paired speech and text $(\mathbf{x}, \mathbf{y})$, we can generate $p_{y_t}$ from autoregressive ASR decoder $Dec_{ASR}$, conditioned to ground-truth text $y_{t-1}$ in the current time-step and encoded speech feature $\mathbf{h}^e$. At the end, the length of probability vector $\mathbf{p}_y$ is fixed to $T$ time-steps.
		\item Conditional generation given previous step model prediction: \\
		Another generation process to decode ASR transcription uses its own prediction to generate probability vector $p_{y_t}$. There are many different generation methods, such as greedy decoding (1-best beam-search) ($\tilde{y_t} = \argmaxA_{c} p_{y_t}[c]$), beam-search, or stochastic sampling ($\tilde{y}_{t} \sim Cat(p_{y_t})$).
	\end{enumerate}
	After the generation process, we obtained probability vector $\mathbf{p}_y$ and applied discretization from continuous probability vector $p_{y_{t}}$ to $\tilde{y}_t$ either by taking the class with the highest probability or sampling from a categorical random variable. After getting a single class to represent the probability vector, we encode it into vector $[0,0,..,1,..,0]$ with one-hot encoding representation and give it to the TTS as the encoder input. The TTS reconstructs Mel-spectrogram $\hat{\mathbf{x}}$ with the teacher-forcing approach. The reconstruction loss is calculated:
	\begin{align}
	\mathcal{L}^{rec}_{TTS} = \frac{1}{S} \sum_{s=1}^{S} (x_s - \hat{x}_s)^2, \label{eq:loss_recon}
	\end{align} where $\hat{x}_s$ is the predicted (or reconstructed) Mel-spectrogram and $x_s$ is the ground-truth spectrogram at $s$-th time-step.
	
	We directly calculated the gradient from the reconstruction loss w.r.t TTS parameters ($\partial \L_{TTS}^{rec} / \partial \theta_{TTS}$) because all the operations inside the TTS module are continuous and differentiable. However, we could not calculate the gradient from the reconstruction loss w.r.t ASR parameters ($\partial \L_{TTS}^{rec} / \partial \theta_{ASR}$) because we have a discretization operation from $p_{y_t} \rightarrow \texttt{onehot}(\tilde{y}_t)$. Therefore, we applied a straight-through estimator to enable the loss from $\L_{TTS}^{rec}$ to pass through discrete variable $\tilde{y}_t$.

	\subsection{Straight-through Argmax} 
	\label{sec:argmax}
	
\begin{figure}[h]
		\centering
		\includegraphics[width=0.55\linewidth]{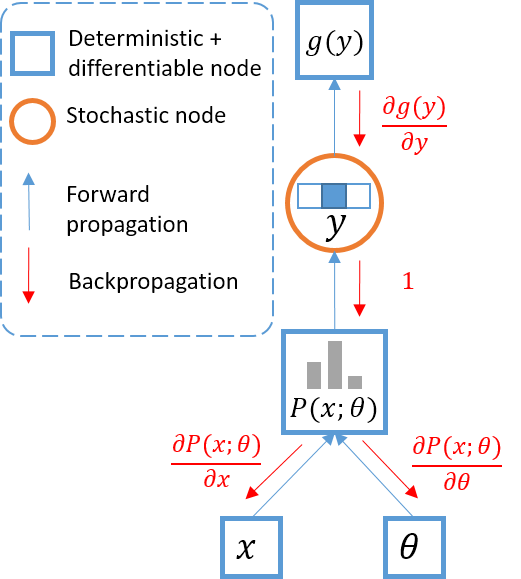}
		
		\caption{\textbf{Straight-through estimator on $\arg max$ function}. Given input $x$ and model parameters $\theta$, we calculate categorical probability mass $P(x; \theta)$ and apply discrete operation $\argmax$. In the backward pass, the gradient from stochastic node $y$ to $P(x; \theta)$, $\partial y / \partial P(x; \theta) \approx \mathbbm{1}$ is approximated by identity.}
		\label{fig:st_estimator}
	\end{figure}

	The straight-through estimator \cite{bengio2013estimating, hinton2012coursera} is a method for estimating or propagating gradients through stochastic discrete variables. Its main idea is to backpropagate through discrete operations (e.g., $\argmaxA_c p_{y_t}[c]$ or sampling $\tilde{y}_t \sim Cat(p_{y_t})$) like an identity function. We describe the forward process and the gradient calculation with a straight-through estimator in Fig.~\ref{fig:st_estimator}.

	In the implementation, we created a function with different forward and backward operations. For $\argmax$ one-hot encoding function, we formulated the forward operation:
	
	\begin{eqnarray}
	\tilde{z}_t = \argmaxA_{c}p_{y_t}[c] \\
	\tilde{y}_t = \texttt{onehot}(\tilde{z}_t).
	\end{eqnarray}
	
	Here we describe $\tilde{y}_t$ as a one-hot encoding vector with the same length as the $p_{y_t}$ vector. When the loss is calculated and the gradients are backpropagated from loss $\mathcal{L}_{TTS}^{rec}$, we formulate the backward operation:
	
	\begin{equation}
	\frac{\partial \tilde{y}_t}{\partial p_{y_t}} \approx \mathbbm{1}. \label{eq:straight_through_identity}
	\end{equation}
	
	Therefore, when we back-propagate the loss from Eq. \ref{eq:loss_recon} with the straight-through estimator approach, we calculate the TTS reconstruction loss gradient w.r.t $\theta_{ASR}$:
	\begin{align}
	\frac{\partial \mathcal{L}_{TTS}^{rec}}{\partial \theta_{ASR}} &=\sum_{t=1}^{T} \frac{\partial \mathcal{L}_{TTS}^{rec}}{\partial \tilde{y_t}} \cdot \frac{\partial {\tilde{y_t}}}{\partial p_{y_t}} \cdot \frac{\partial p_{y_t}}{\partial \theta_{ASR}} \\
	&\approx \sum_{t=1}^{T} \frac{\partial \mathcal{L}_{TTS}^{rec}}{\partial \tilde{y_t}} \cdot \mathbbm{1} \cdot \frac{\partial p_{y_t}}{\partial \theta_{ASR}}.
	\end{align}

	\subsection{Straight-through Gumbel Softmax} 
	\label{sec:gumbel_softmax}
	
	Besides taking $\argmax$ class from probability vector $p_{y_t}$, we also generated a one-hot encoding by sampling with the Gumbel-Softmax distribution \cite{evjang2018gumbel, maddison2016concrete}. Gumbel-Softmax is a continuous distribution that approximates categorical samples, and the gradients  can be calculated with a reparameterization trick. For Gumbel-Softmax, we replaced the softmax formula for calculating $p_{y_t}$ (Eq.~\ref{eq:softmax}):
	
	\begin{equation}
	p_{y_t}[c] = \frac{\exp((h^d_t[c] + g_c) /\tau)}{\sum_{i=1}^{C} \exp((h^d_t[i] + g_i) / \tau)}, \quad \forall c \in [1..C]. \label{eq:gumbel_softmax}
	\end{equation} 
	
	where $g_1,..,g_C$ are i.i.d samples drawn from Gumbel(0, 1) and $\tau$ is the temperature. We sample $g_i$ by drawing samples from the uniform distribution:
	
	\begin{eqnarray}
	&u_c \sim \texttt{Uniform}(0, 1) \\
	&g_c = -\log(-\log(u_c)), \quad \forall c \in [1..C].
	\end{eqnarray}
	
	To generate a one-hot encoding, we define our forward operation:
	
	\begin{eqnarray}
	&\tilde{z}_t \sim Categorical(p_{y_t}[1], p_{y_t}[2], ..., p_{y_t}[C]) \label{eq:sampling_cat} \\
	&\tilde{y}_t = \texttt{onehot}(\tilde{z}_t).
	\end{eqnarray}
	
	At the backpropagation time, we use the same straight-through estimator (Eq.~\ref{eq:straight_through_identity}) to allow the gradients to flow through the discrete sampling operation from Eq.~\ref{eq:sampling_cat}.

	\subsection{Combined Loss for ASR}
	
	Our final loss function for ASR is a combination from negative likelihood (Eq.~\ref{eq:loss_asr}) and TTS reconstruction loss (Eq.~\ref{eq:loss_recon}) by sum operation:
	
	\begin{equation}
	\L_{ASR}^F = \L_{ASR} + \L_{TTS}^{rec}.
	\end{equation}
	
\begin{figure}[h]
		\centering
		\includegraphics[width=0.65\linewidth]{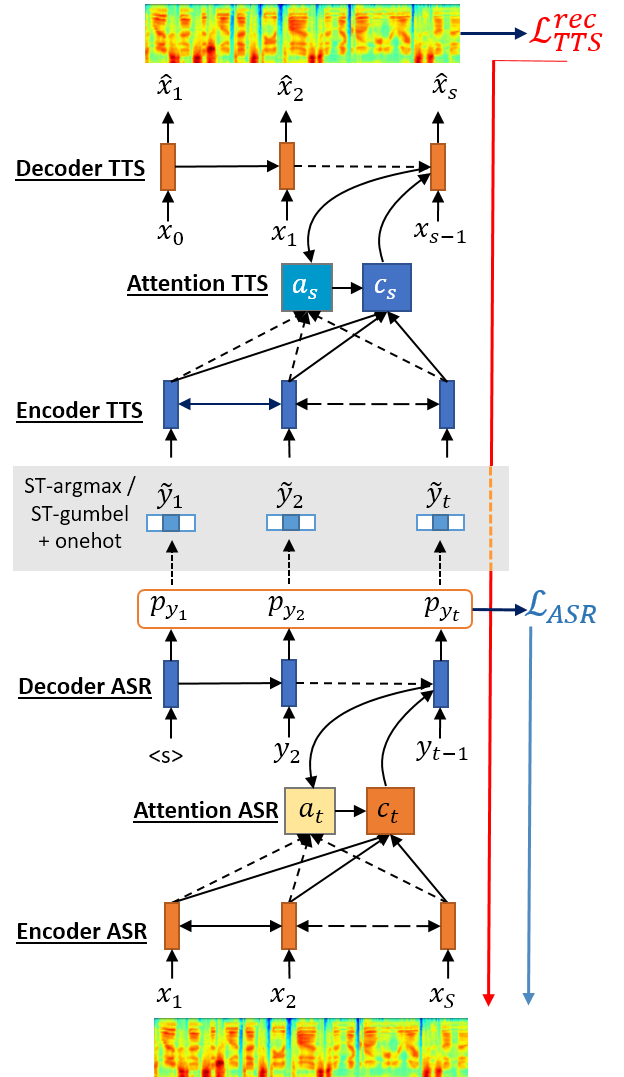}
		
		\caption{Given speech feature $\mathbf {x}$, ASR generates a sequence of probability $\mathbf{p}_y = [p_{y_1}, p_{y_2},...,p_{y_T}]$. If we have a ground-truth transcription, we can calculate $\L_{ASR}$ (Eq.~\ref{eq:loss_asr}).  TTS module generates speech features, and we calculate reconstruction loss $\L_{TTS}^{rec}$ (Eq.~\ref{eq:loss_recon}). After that, the gradients based on $\L_{ASR}$ are  propagated through the ASR module, and the gradients based on $\L_{TTS}^{rec}$ are propagated through the TTS and ASR modules by a straight-through estimator.}
		\label{fig:speech_chain_asr_tts}
	\end{figure}

	To summarize our explanation in this section, we provide an illustration in Fig.~\ref{fig:speech_chain_asr_tts} that explains how sub-losses $\L_{ASR}$ and $\L_{TTS}^{rec}$ are backpropagated to the rest of the ASR and TTS modules.
	
	\section{Sequence-to-Sequence Model for ASR} \label{sec:intro_asr}
	
	A sequence-to-sequence model is a neural network that directly models conditional probability $p({y}|{x})$, where $\mathbf{x} = [x_1, ..., x_S]$ is the sequence of the (framed) speech features with length $S$ and $\mathbf{y} = [y_1, ..., y_T]$ is the label’s sequence with length $T$.
	
	The encoder task processes input sequence ${x}$ and generating  representative information ${h^e} = [h^e_1, ...,h^e_S]$ for the decoder. The attention module is an extension scheme that assists the decoder to find relevant information on the encoder side based on the current decoder hidden states \cite{bahdanau2014neural, chan2016listen}. Attention modules produce context information $c_t$ at time $t$ based on the encoder and decoder hidden states:
	
	\begin{align}
	c_t &= \sum_{s=1}^{S} a_t(s) * h^e_s \\
	a_t(s) &= \text{Align}({h^e_s}, h^d_t) \nonumber \\
	&= \frac{\exp(\text{Score}(h^e_s, h^d_t))}{\sum_{s=1}^{S}\exp(\text{Score}(h^e_s, h^d_t))}.
	\end{align}
	
	There are several variations for score functions \cite{luong2015effective} such as $Score(h_s^e, h_t^d)$:
	%    
	%\begin{align}
	%\text{Score}(h_s^e, h_t^d) =
	%\begin{cases}
	%\langle h_s^e, h_t^d\rangle, & \text{dot product}  \\
	%h_s^{e\intercal} W_{s} h_t^d, & \text{bilinear}  \\
	%V_s^{\intercal} \tanh(W_{s} [h_s^e, h_t^d]), & \text{MLP} \label{eq:mlpscore}  \\
	%\end{cases}
	%\end{align} 
	%    
	dot product ($\langle h_s^e, h_t^d \rangle$), bilinear ($h_s^{e\intercal} W_s h_t^d$), where $\text{score}:(\mathbb{R}^M \times \mathbb{R}^N) \rightarrow \mathbb{R}$, $M$ is the number of hidden units for the encoder and $N$ is the number of hidden units for the decoder.
	Finally, the decoder task predicts target sequence probability $p_{y_t}$ at time $t$ based on previous output and context information $c_t$. The loss function for ASR can be formulated:
	
	\begin{equation}
	\L_{ASR} = -\frac{1}{T}\sum_{t=1}^{T}\sum_{c=1}^{C}\mathbbm{1}(y_t=c)*\log{p_{y_t}}[c], \label{eq:loss_asr}
	\end{equation} 
	
	where $C$ is the number of output classes.
	Input ${x}$ for the speech recognition tasks is a sequence of feature vectors like a log Mel-scale spectrogram. Therefore, ${x} \in \mathbb{R}^{S \times D}$, where D is the number of features and S is the total frame length for an utterance. Output ${y}$, which is a speech transcription sequence, can be either a phoneme or a grapheme (character) sequence.

	\section{Sequence-to-Sequence Model for TTS} \label{sec:intro_tts}

	Speech synthesis can be viewed as a sequence-to-sequence task where a model generates speech given a sentence. We directly model the conditional probability $p(x|y)$ with a sequence-to-sequence model, where $y=[y_1,...,y_T]$ is the sequence of characters with length $T$ and $x=[x_1,...,x_S]$ is the sequence of (framed) speech features with length $S$. From the sequence-to-sequence ASR model perspective, TTS is the reverse case where the model reconstructs the original speech given the text.
	
	In this work, our core architecture is based on Tacotron \cite{wang2017tacotron} with several structural modifications \cite{tjandra2018machinespeech}.
	%Fig.~\ref{fig:seq2seq_tts} illustrates our modified Tacotron.
	The main difference between our modified Tacotron and the default Tacotron is that we added an additional speaker embedding projection layer into our decoder to enable multispeaker training and generation. We also have an additional output layer to generate binary prediction $b_s \in [0, 1]$ (1 if the $s$-th frame is the end of speech, otherwise 0).
	
	For training the TTS model, we used the following loss function:
	
	\begin{equation}
	\begin{split}
	\L_{TTS} = & \frac{1}{S}\sum_{s=1}^{S} (x_s^M - \hat{x}_s^M)^2 + (x_s^R - \hat{x}_s^R)^2 \\
	& - (b_s \log(\hat{b}_s) + (1-b_s) \log(1-\hat{b}_s)),
	\end{split}
	\end{equation} 
	
	where $\hat{x}^M, \hat{x}^R, and \hat{b}$ are the predicted log Mel-scale spectrogram, the log magnitude spectrogram, and the end-of-frame probability, and $x^M, x^R, b$ is the ground-truth. In the decoding process, we use the Griffin-Lim algorithm \cite{griffin1984signal} to iteratively estimate the phase spectrogram and reconstruct the signal with inverse STFT.

	\section{Experiment} \label{sec:result}
	
	\subsection{Dataset}
	
	We evaluated the performance of our proposed method on the Wall Street Journal dataset \cite{paul1992design}. Our settings for the training, development, and test sets are the same as the Kaldi s5 recipe \cite{povey11asru}. We trained our model with WSJ-SI284 data. Our validation set was dev\_93, and our test set was eval\_92.
	
	We used the character sequence as our decoder target and followed the preprocessing steps proposed by a previous work \cite{hannun2014first}. The text from all the utterances was mapped into a 32-character set: 26 (a-z) letters of the alphabet, apostrophes, periods, dashes, space, noise, and ``eos.'' In all the experiments, we extracted the 40 dims + $\Delta$ + $\Delta\Delta$ (total 120 dimensions) log Mel-spectrogram features from our speech and normalized every dimension into zero mean and unit variance.

	\subsection{Model Details}
	
	For the ASR model, we used a standard sequence-to-sequence model with an attention module (Section~\ref{sec:intro_asr}). On the encoder sides, the input log Mel-spectrogram features were processed by three bidirectional LSTMs (Bi-LSTM) with 256 hidden units for each LSTM: a total of 512 hidden units for the Bi-LSTM. To reduce the memory consumption and processing time, we used hierarchical sub-sampling \cite{graves2012supervised, bahdanau2016end} on all three Bi-LSTM layers and reduced the sequence length by a factor of eight. On the decoder sides, we projected one-hot encoding from the previous character into a 256-dims continuous vector with an embedding matrix, followed by one unidirectional LSTM with 512 hidden units. For the attention module, we used the content-based attention + multiscale alignment (denoted as ``Att MLP-MA") \cite{tjandra2018multi} with a 1-history size. In the decoding phase, the transcription was generated by beam-search decoding (size=5), and we normalized the log-likelihood score by dividing it with its own length to prevent the decoder from favoring shorter transcriptions. We did not use any language model or lexicon dictionary in this work. In the training stage, we tried ST-argmax (Section~\ref{sec:argmax}) and ST-gumbel softmax (Section~\ref{sec:gumbel_softmax}). We also tried both teacher-forcing and greedy decoding to generate ASR probability vectors $\mathbf{p}_{y}$. For each scenario, we treated temperature $\tau = [0.25, 0.5, 1, 2]$ as our hyperparameter and searched for the best temperature based on the CER (character error rate) on the development set.
	
	For the TTS model, we used the TTS explained in Section~\ref{sec:intro_tts}. The hyperparameters for the basic structure are generally the same as those for the original Tacotron, except we replaced ReLU with the LReLU function. For the CBHG module, we used $K=8$ filter banks instead of 16 to reduce the GPU memory consumption. For the decoder sides, we deployed two LSTMs instead of a GRU with 256 hidden units. For each time-step, our model generated four consecutive frames to reduce the number of steps in the decoding process.

	\subsection{Experiment Result}
	
	\begin{table}[]
		\centering
		\footnotesize
		\label{tbl:result_asr}
		\caption{ASR experiment result on WSJ dataset test\_eval92.}
		\begin{tabular}{|l|l|l|c|}
			\hline
			\multicolumn{4}{|c|}{\cellcolor[HTML]{EFEFEF}\textbf{Baseline ($\L_{ASR}$)}}                                                    \\ \hline
			\multicolumn{3}{|c|}{\textbf{Model}}                                                                        & \textbf{CER (\%)} \\ \hline
			\multicolumn{3}{|l|}{Att MLP \cite{kim2017joint}}                                                           & 11.08             \\ \hline
			\multicolumn{3}{|l|}{Att MLP + Location \cite{kim2017joint}}                                                & 8.17              \\ \hline
			\multicolumn{3}{|l|}{Att MLP \cite{tjandra2017attention}}                                                   & 7.12              \\ \hline
			\multicolumn{3}{|l|}{Att MLP-MA (ours) \cite{tjandra2018multi}}                                                    & 6.43              \\ \hline
			\multicolumn{4}{|c|}{\cellcolor[HTML]{EFEFEF}\textbf{Proposed ($\L_{ASR} + \L_{TTS}^{rec}$)}}                                   \\ \hline
			\multicolumn{1}{|c|}{\textbf{Model}} & \multicolumn{1}{c|}{\textbf{Generation}} & \multicolumn{1}{c|}{\textbf{ST}} & \textbf{CER (\%)} \\ \hline
			Att MLP-MA                           &                                   & argmax                           & 5.75              \\ \cline{1-1} \cline{3-4}
			Att MLP-MA                           & \multirow{-2}{*}{Teacher-forcing} & gumbel                           & \textbf{5.7}               \\ \hline
			Att MLP-MA                           &                                   & argmax                           & 5.84              \\ \cline{1-1} \cline{3-4}
			Att MLP-MA                           & \multirow{-2}{*}{Greedy}          & gumbel                           & 5.88              \\ \hline
		\end{tabular}
	\end{table}
	
	For our baseline, we trained an encoder-decoder with MLP + multiscale alignment with a 1-history size \cite{tjandra2018multi}. We also added several published results to our baseline. All of the baseline models were trained by minimizing negative log-likelihood $\L_{ASR}$ (Eq.~\ref{eq:loss_asr}).
	
	All the models in the proposed section were trained with a combination from two losses: $\L_{ASR} + \L_{TTS}^{rec}$, and the ASR parameters were updated based on the gradient from the sum of the two losses. We have four different scenarios, most of which provide significant improvement compared to the baseline model that is only trained on $\L_{ASR}$ loss. With teacher-forcing and sampling from Gumbel-softmax, we obtained 11\% relative improvement compared to our best baseline Att MLP-MA.

	\section{Related Works}
	
	Approaches that utilize end-to-end feedback learning from source-to-target and vice-versa remain scant. Senrich et al. \cite{sennrich2016improving} improved the NMT performance by back-translation on a monolingual dataset. Semi-supervised learning for NMT called dual learning \cite{he2016dual} was also proposed by combining reconstruction loss and language model reward. However, the feedback gradient provided by the reconstruction loss only limited the closest module to the loss. One primary reason is that the nature of text modalities is represented by discrete variables. Our previous speech chain paper \cite{tjandra2017speechchain, tjandra2018machinespeech} focused on utilizing the closed-loop between ASR and TTS as a semi-supervised learning method. If one of the modalities of data is missing, we can generate a pseudo-pair and train one of the models by reconstruction loss. But, as we described earlier, the study also limit the back-propagation to the closest module due to similar reason that the output of the ASR is discrete tokens. In contrast, in this paper, we successfully address the problem using a straight-through estimator to predict the gradient through discrete variables.

	\section{Conclusions} \label{sec:conclusion}
	
	We introduced a different perspective from a speech chain mechanism.  We trained our ASR module by adding feedback from the TTS reconstruction loss. However, the ASR output is not differentiable because of the transcription generated by the discretization process. To address this problem, we used a straight-through estimator to enable the gradient from the TTS module to flow through discrete variables . We tried various scenarios with different decoding and discretization processes. From our experimental results, with teacher-forcing and sampling from Gumbel-Softmax, we improved the ASR performances by 11\% relative CER reduction compared to our baseline.

	\section{Acknowledgment}
	
	Part of this work was supported by JSPS KAKENHI Grant Numbers JP17H06101 and JP17K00237.
	\bibliographystyle{ieeetr}
	\bibliography{refs}
\end{document}